\documentclass[10pt,twocolumn,letterpaper]{article}

\usepackage{cvpr}      

\usepackage[dvipsnames]{xcolor}

\usepackage{amsmath,amsfonts,bm}

\def\eqref#1{equation~\ref{#1}}

\def\1{\bm{1}}

\def\eps{{\epsilon}}

\def\rmE{{\mathbf{E}}}

\def\rmH{{\mathbf{H}}}

\def\rmU{{\mathbf{U}}}

\def\ve{{\bm{e}}}

\def\vh{{\bm{h}}}

\def\vu{{\bm{u}}}

\def\vx{{\bm{x}}}

\def\vz{{\bm{z}}}

\DeclareMathAlphabet{\mathsfit}{\encodingdefault}{\sfdefault}{m}{sl}
\SetMathAlphabet{\mathsfit}{bold}{\encodingdefault}{\sfdefault}{bx}{n}

\def\gA{{\mathcal{A}}}

\def\sH{{\mathbb{H}}}
\def\sI{{\mathbb{I}}}

\def\sO{{\mathbb{O}}}

\def\sR{{\mathbb{R}}}

\DeclareMathOperator*{\argmin}{arg\,min}

\usepackage{wrapfig}

\definecolor{cvprblue}{rgb}{0.21,0.49,0.74}
\usepackage[pagebackref,breaklinks,colorlinks,citecolor=cvprblue]{hyperref}

\def\paperID{*****} 
\def\confName{CVPR}
\def\confYear{2024}

\title{The Deep Equilibrium Algorithmic Reasoner}

\author{%
Dobrik Georgiev\\
University of Cambridge\\
{\tt\small dgg30@cam.ac.uk} \\
\and
Pietro Liò \\
University of Cambridge \\
{\tt\small pl219@cam.ac.uk} \\
\and
Davide Buffelli\\
MediaTek Research \\
{\tt\small davide.buffelli@mtkresearch.com} \\
}

\begin{document}
\maketitle
\begin{abstract}
Recent work on neural algorithmic reasoning has demonstrated that graph neural
networks (GNNs) could learn to execute classical algorithms. Doing so, however,
has always used a recurrent architecture, where each iteration of the GNN
aligns with an algorithm's iteration. Since an algorithm's solution is often an
equilibrium, we conjecture and empirically validate that one can train
a network to solve algorithmic problems by directly finding the equilibrium.
Note that this does not require matching each GNN iteration with a step of the
algorithm.
\end{abstract}

\section{Introduction}

Neural Algorithmic Reasoning (a.k.a. NAR; \citealp{velickovic2021neural})
models are a class of neural networks (usually graph neural networks, a.k.a.
GNNs) that learn to imitate classical algorithms \citep{ibarz2022generalist}.
One of the key reasons that this performance is achievable is
\textit{alignment} \citep{xu2020what}: GNNs that align better to the target
algorithm achieve better generalisation. This alignment game has led to
a sequence of exciting research -- from aligning the architecture with
iterative algorithms \citep{tang2020towards} to proving that ``graph neural
networks are dynamic programmers'' \citep{dudzik2022graph}.\footnote{We refer
the reader to \citet{velickovic2023NARGradient} for a more comprehensive
overview of NAR, the ``alignment game'' and examples of why learning to
execute algorithms is useful in real-world practice.}

Notably, the aforementioned papers focus on aligning the computation of the GNN
with an algorithm or a specific algorithm class (e.g. dynamic programming), but
ignore the properties \emph{at the time of algorithm termination}. For a number
of algorithms (e.g. sorting, shortest-path and dynamic programming) in the
CLRS-30 algorithmic benchmark \citep{velivckovic2022clrs} once the optimal
solution is found, further algorithm iterations will not change the algorithm's
output prediction values. For example, in shortest-paths algorithms \citep[e.g.
ones found in][]{cormen2009tsp} making additional iterations would not alter
the optimality of the shortest paths' distances found. We will call such state
an \emph{equilibrium} -- additional applications of a function (an algorithm's
iteration) to the state leave it unchanged. There are other examples of
algorithms in the benchmark that have equilibriums at their solved state:
Kruskal's \citep{kruskal1956shortest} minimum spanning tree
algorithm\footnote{Once the tree is built, iterating over the rest of the
edges, if any remain, will not change the solution.}, sorting algorithms, etc.

In this paper, we show that:
\begin{enumerate}
    \item We investigate a novel approach to learning algorithms by identifying
        the equilibrium point of the GNN equation.
    \item By aligning the NAR models to the equilibrium property discussed in
        the previous paragraph we can improve the model accuracy.
    \item By removing the requirement that one step of the GNN $\leftrightarrow$ one
        step of the algorithm and by finding equilibrium points it is possible to
        reduce the required number of GNN iterations.
\end{enumerate}

The rest of the paper is organised as follows: \cref{sec:background} provides
a short background on NAR and deep equilibrium models (DEQs; the approach we
used), \cref{sec:DER} outlines our model in more detail, presenting
theoretically why we can express NAR with DEQ, \cref{sec:experimental} gives
information on our experimental setup and presents the results.

\section{Background}\label{sec:background}

\subsection{Algorithmic Reasoning}

Let $A: \sI_A \to \sO_A$ be an algorithm, acting
on some input $\vx \in \sI_A$, producing an output $A(\vx) \in \sO_A$ and let
$\sI_A$/$\sO_A$ be the set of possible inputs/outputs $A$ can read/return. In
algorithmic reasoning, we aim to learn a function $\gA: \sI_A \to \sO_A$, such
that $\gA \approx A$.  Importantly, we will not be learning simply an
input-output mapping, but we will aim to align to the algorithm $A$'s
trajectory. The alignment is often achieved through direct
supervision\footnote{Recent research \citep{bevilacqua2023neural,
rodionov2023neural} has shown that alternative, causality-inspired, ways of
alignment also exist.} on the intermediate states of the algorithm. To capture
the execution of $A$ on an input $x$ we can represent it as
\begin{align}
    \bar{\vh}_0 &= \textsc{Preprocess}(\vx)\\
    \bar{\vh}_\tau &= \underbrace{A_t(A_t(\dots A_t}_\text{$\tau$ times}(\bar{\vh}_0)\dots))\label{eq:algorollout}\\
    A(\vx) &= \textsc{Postprocess}(\bar{\vh}_\tau) 
\end{align}
where \textsc{Preprocess} and \textsc{Postprocess} are some simple pre- and
post-processing (e.g.initialising auxiliary variables or returning the correct
variable), $\bar{\vh}_\tau\in \sH_A$ is $A$'s internal (\textbf{h}idden) state
and $A_t$ is a subroutine (or a set of) that is executed at each step. It is
therefore no surprise that the encode-process-decode architecture
\citep{hamrick2018relational} is the de-facto choice when it comes to NAR.
Thus, the architecture can be neatly represented as a composition of three
learnable components: $\gA= g_\gA \circ P \circ f_\gA$, where $g_\gA: \sI_A \to
\sR^d$ and $f_\gA: \sR^d \to \sO_A$ are the encoder and decoder function
respectively (usually linear projections) and $P: \sR^d \to \sR^d$ is
a processor that mimics the rollout (\cref{eq:algorollout}) of $A$. The
processor is often modelled as a message-passing GNN with the message function
containing nonlinearities (e.g. ReLU).

\paragraph{CLRS-30}The \emph{CLRS-30} benchmark \citep{velivckovic2022clrs}
includes 30 iconic algorithms from the \textit{Introduction to Algorithms}
textbook \citep{CLRS}. Each data instance for an algorithm $A$ is a graph
annotated with features from different algorithm stages (\emph{input},
\emph{output}, and \emph{hint}), each associated with a location (\emph{node},
\emph{edge}, and \emph{graph}). Hints contain time series data representing the
algorithm rollout and include a temporal dimension often used to infer the
number of steps $\tau$. Features in CLRS-30 have various datatypes with
associated losses for training. The test split, designed for assessing
out-of-distribution (OOD) generalization, comprises graphs four times larger
than the training set. For more details, see \citet{velivckovic2022clrs}.

\subsection{Deep Equilibrium Models}

Deep equilibrium models \citep[DEQs][]{bai2019deep} are a class of
implicit neural networks \citep{ghaoui2019implicit}. The functions modelled
with DEQs are of the form:
\begin{align}
    \vz^*=f_\theta(\vz^*, \vx)\label{eq:fixedpoint}
\end{align}
where $\vx$ is input, $f_\theta$ is a function parametrised by $\theta$ (e.g.
a neural network) and $\vz^*$ is the output. $\vz^*$ is an equilibrium point to
the eventual output value of an infinite depth network where each layer's
weights are shared, i.e.  $f_\theta^{[i]}=f_\theta$. By re-expressing
(\ref{eq:fixedpoint}) as $g_\theta=f_\theta(\vz^*, \vx)-\vz^*$ DEQs allow us to
find the fixed point $\vz^*$ via any black-box root-finding method
\citep[e.g.][]{broyden1965aclass, anderson1965iterative}, without the actual
need of unrolling the equation until convergence. DEQs also integrate with
backpropagation -- gradient $\partial \mathcal{L}/\partial \theta$ could be
calculated using the Implicit Function Theorem (cf. \citealp{bai2019deep}) and
no intermediate state has to be stored, giving us constant memory cost of
gradient computation regardless of the number of iterations until convergence.

Another nice property DEQs possess is that they are independent of the
choice of $f_\theta$. Ultimately, this is what allows us to integrate them
with ease with algorithmic reasoner architectures. Other implicit GNN models
\citep[e.g][]{gu2020implicit} do not provide the flexibility to incorporate edge
features or use a \texttt{max} aggregator.

\section{The Deep Equilibrium Algorithmic Reasoner}\label{sec:DER}

\subsection{Motivation} The previous section has hinted at the existence of an
alignment between equilibrium models and algorithms. This was also observed
empirically by \citet{mirjanic2023latent}, who showed the presence of 
attractor states in the latent space trajectories of the algorithm: graphs of
similar executions tend to cluster together and do not diverge. This further
motivates the usage of DEQs for NAR since finding the equilibrium of
$f_\theta$ reliably and efficiently requires the function to be stable.

\subsection{Implementation}\label{subsec:impl}
We implement our processor as a PGN architecture
\citep{velickovic2020pointer} with a gating mechanism as in
\citet{ibarz2022generalist}:
\begin{align}
        \mathbf{z}_i^{(t)} &= \mathbf{u}_i \Vert \mathbf{h}_i^{(t-1)}\\
    \label{eq:processor_start}
        \mathbf{m}_i^{(t)} &= \max_{j \in \mathcal{N}_i} P_m\left(\mathbf{z}_i^{(t)}, \mathbf{z}_j^{(t)}, \mathbf{e}_{ij} \right)\\
        \mathbf{\hat{h}}_i^{(t)} &= P_r\left(\mathbf{z}_i^{(t)}, \mathbf{m}_i^{(t)}\right) \\
        \mathbf{g}_i^{(t)} &= P_g(\mathbf{z}_i^{(t)}, \mathbf{m}_i^{(t)})\\
        \mathbf{h}_i^{(t)} &= \mathbf{g}_i^{(t)} \odot \mathbf{\hat{h}}_i^{(t)} + (1-\mathbf{g}_i^{(t)}) \odot \mathbf{\hat{h}}_i^{(t-1)}
    \label{eq:processor_end}
\end{align}
\begin{figure}
    \centering
    \includegraphics[width=.8\linewidth]{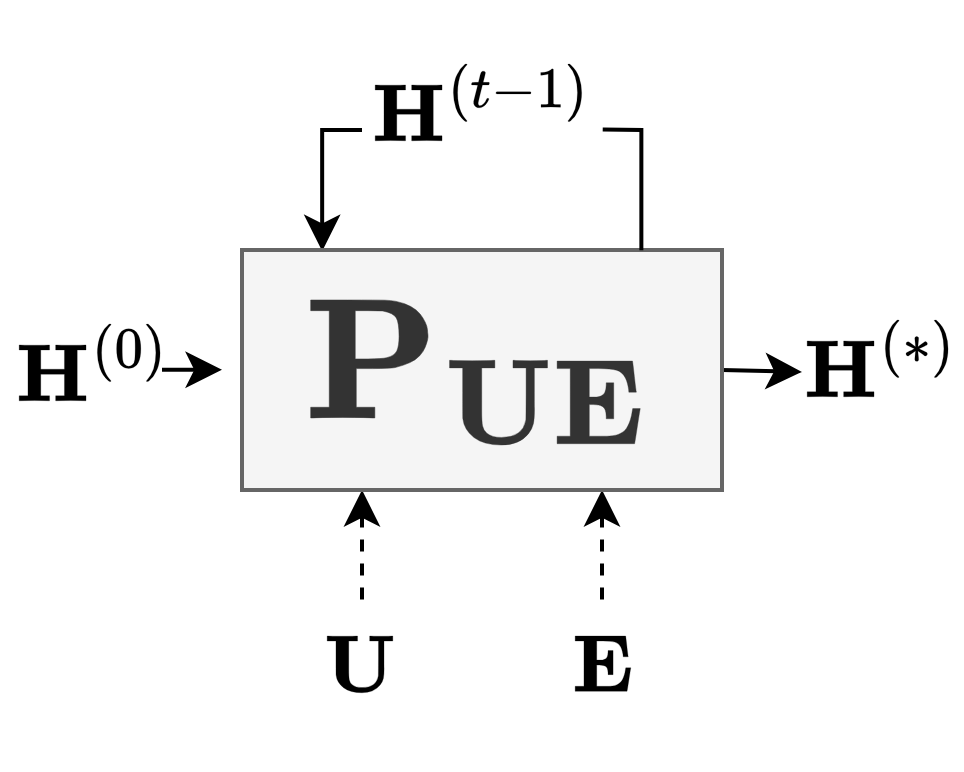}
    \caption{
        The deep equilibrium algorithmic reasoner. Dashed arrows represent constant
        variables.
    }
\end{figure}
where $\mathbf{u}_i$ and $e_{ij}$ are the node and edge input
features\footnote{Obtained by linearly encoding algorithm's inputs},
$\vh_i^{(t)}$ is the latent state of node $i$ at timestep $t$ (with
$\vh_i^{(0)}=\mathbf{0}$), $\Vert$ and $\odot$ denotes concatenation and
elementwise multiplication. $P_m$, $P_g$ and $P_r$ are the processor's message,
gating and readout functions. $P_m$ is parametrised as an MLP, the rest are
linear projection layers. Eqs
(\ref{eq:processor_start})-(\ref{eq:processor_end}) can also be viewed as
a processor function $P(\rmH^{(t-1)}, \rmU, \rmE)$ taking node ($\rmU$), edge
($\rmE$) and previous latent ($\rmH^{(t-1)}$) features. Noting that $\vu_i$ and
$\ve_{ij}$ do not change between iterations for all $i$ and $j$, the fixed
point of the rollout of Eqs (\ref{eq:processor_start})-(\ref{eq:processor_end})
can be expressed as:
\begin{align}
    \begin{split}
        \rmH^{(*)}=P_{\rmU\rmE}(\rmH^{(*)})
    \end{split}\label{eq:DER}
\end{align}
where $P_{\rmU\rmE}(\rmH^{(*)})=P(\rmH^{(*)}, \rmU, \rmE)$. We want to
emphasise that for a given problem instance $\rmU$ and $\rmE$ do not change
during the search for equilibrium. The above \cref{eq:DER} matches the
signature of \cref{eq:fixedpoint}, is solved via root-finding (as if it is
$f_\theta$ of a DEQ) and is what we will call the \emph{deep equilibrium
algorithmic reasoner} (DEAR) in our experiments. The rest of NAR
architecture, i.e.  encoders and decoders are implemented as in
\citet{ibarz2022generalist}.

\paragraph{A note on underreaching} In the DEQ implementation we
used\footnote{\texttt{torchdeq} \citep{torchdeq}, MIT License} \emph{one step
of the solver calls the GNN processor exactly once}. It is in theory
possible that the number of solver iterations needed to find equilibrium is
less than the diameter of the graph, resulting in underreaching: the
information from one node (e.g.  starting node in a shortest-path
algorithm) cannot reach a target node. In our experiments, however, the
solver either needed more iterations than the ground-truth or, if it
needed less, was on data instances that are represented as full-graphs (sorting in CLRS).

\paragraph{A note on oversmoothing} A similar argument can be made, but in the
opposite direction -- what happens if the solver has to use more
calls/iterations and does this lead to oversmoothing? First, we did not
empirically observe such an issue -- if all $\vh^{(*)}_i$ are equivalent, we
would be decoding the same outputs across all nodes/edges and accuracy/loss
will suffer. Second, our GNN model makes use of gating (cf.
\cref{eq:processor_end}). Although less intricate than works targeting
oversmoothing, according to a recent survey on GNN \citep{rusch2023survey}
gating is one of the strategies to avoid it.

\paragraph{A note on hints} As we already mentioned, in Eqs
(\ref{eq:processor_start})-(\ref{eq:processor_end}) the node and edge features
are the same across timesteps and this is due to the fact we do not make any
use of hints (i.e. predicting intermediate algorithm state). First, although it
may seem counterintuitive, it has been shown that a NAR model can successfully
generalise, even when trained to only predict the correct output
\citep{bevilacqua2023neural}. Second, the fact that the solver uses the GNN
exactly once per call \emph{does not imply that one step of the solver
corresponds to one iteration of the algorithm}. Adding hint-based loss,
be it directly supervising on the outputs or performing contrastive learning,
based on hint values as in \citet{bevilacqua2023neural}, is ambiguous as there
is no one-to-one correspondence of solver iteration and step of the algorithm
and is left for future work.

\section{Evaluation}\label{sec:experimental}

\subsection{Setup} For each algorithm we generated $10^5$/100/100-sized
training/validation/test datasets. Training samples sizes vary between 8 and 16 elements
(uniformly randomly chosen), validation and test samples are of size 16 and 64
respectively. For requiring graphs as input (non-sorting tasks in our
experiments) we generate Erdős–Rényi graphs \citep{erdos1960evolution} with
edge probabilities $p$ uniformly randomly sampled from the interval $[0.1,
0.9]$, with increments of $0.1$. We obtained the ground truth execution
trajectories and targets using the CLRS-30
implementation.\footnote{\url{https://github.com/google-deepmind/clrs/},
Apache-2.0 License}

In our experiments the models have a latent dimensionality of 128, the batch
size is 32, learning rate is $3\times10^{-4}$ and we use the Adam optimizer
\citep{kingma2015adam}. We train our algorithmic reasoners for 100 epochs,
choosing the model with the lowest validation loss for testing. Each task is
independently learned, minimising the output loss (losses depend on the
algorithm, cf.  CLRS-30). We did not include hint trajectories in the loss
computation, since as already mentioned in \S\ref{subsec:impl} adding
supervision on hints is ambiguous.

\begin{figure}
    \includegraphics[width=.9\linewidth]{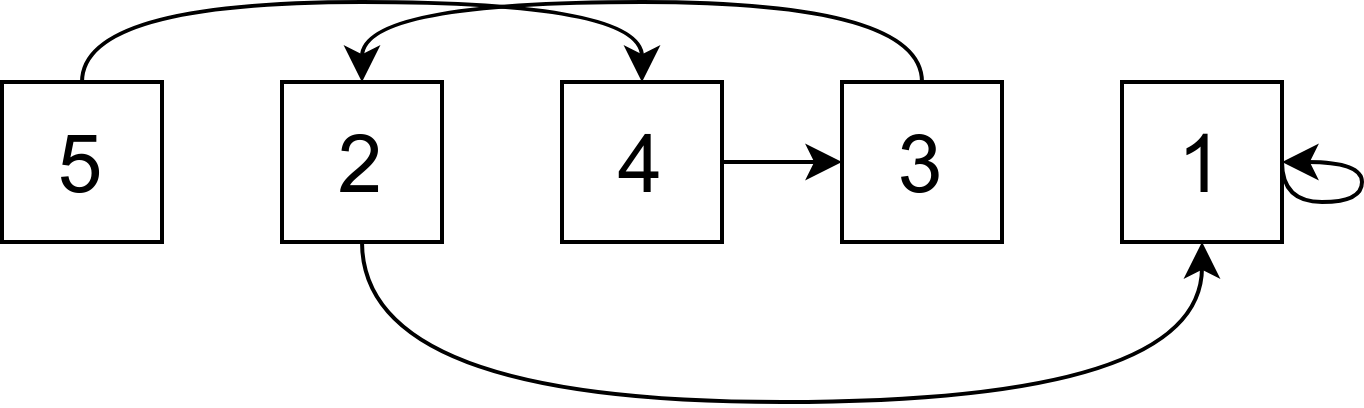}
    \caption{An example input list and predecessors indicating order. Source: \citet{ibarz2022generalist}}\label{fig:example}
\end{figure}
\newcommand{\specialcell}[2][l]{%
  \begin{tabular}{@{}#1@{}}#2\end{tabular}}

\begin{table*}[t]
\centering
\caption{
Performance metrics for different algorithms and models. Note that for DEAR
models the type of sorting algorithm should not matter as the output target is
always the same.
}
\label{tab:OHDEAR}
\begin{tabular}{lcccc}
    \toprule
     & \multicolumn{4}{c}{Algorighm} \\
    \cmidrule(r){2-5}
    \textbf{Model} & Bellman-Ford & Floyd-Warshall & SCC & (Insertion) Sort \\
    \midrule
    \specialcell{\textbf{NAR\footnotemark}\\\footnotesize(Triplet-MPNN)} & $93.26 \pm 0.04\%$ & $40.80 \pm 2.90\%$ & $57.63 \pm 0.68\%$ & $77.29 \pm 7.42\%$ \\
    \midrule
    \textbf{NAR\footnotemark} & $97.03 \pm 0.10\%$ & $52.58 \pm 1.08\%$ & $40.82 \pm 0.75\%$ & $63.20 \pm 11.32\%$ \\
    \specialcell{\textbf{NAR\footnotemark[8]}\\(\footnotesize w/ jac. reg)} & $97.02 \pm 0.08\%$ & $53.34 \pm 1.51\%$ & $39.95 \pm 1.12\%$ & $52.44 \pm 10.07\%$ \\
    \midrule
    \textbf{DEAR} & $97.50 \pm 0.29\%$ & $52.79 \pm 0.32\%$ & $42.04 \pm 0.93\%$ & $85.29 \pm 6.13\%$ \\
    \specialcell[l]{\textbf{DEAR}\\(\footnotesize rel.\ tol.\\\&\footnotesize lower $\eps$)} & $96.78 \pm 0.43\%$ & $52.39 \pm 1.34\%$ & $43.38 \pm 0.39\%$ & $86.93 \pm 3.87\%$ \\
    \bottomrule
\end{tabular}
\end{table*}
The performance metric we measure is the out-of-distribution accuracy. The
definition of accuracy varies between algorithms and is based on the
specification of the algorithm itself (see \citet{ibarz2022generalist} and
CLRS-30 for all possible accuracy metrics). For the four algorithms we evaluate
on, the accuracy metric is going to be the node/edge pointer accuracy:
\begin{itemize}
    \item Bellman-Ford -- predecessor node pointer accuracy -- for each node
        $u$ predict the immediate predecessor on the shortest-path from the
        starting node to $u$
    \item Floyd-Warshall -- predecessor \emph{edge} pointer accuracy -- similar
        to the above. For each edge $(i,j)\in\mathcal{E}$ predict the node $k$,
        such that $k=\argmin_{k\in \mathcal{V}} dist[i][k]+dist[k][j]$
    \item Strongly connected components -- node pointer accuracy -- for
        each node predict a pointer to the immediate predecessor in the
        same component. If more than one such predecessors exist, chose
        the one with the lowest index.
    \item Insertion sort -- node pointer accuracy -- the sorting array iq
        represented by the vector of pointers $\mathbf{\pi}$. $\pi_i$ points
        the the array element before $i$ in the sorted sequence. E.g. in the
        example (\cref{fig:example}) if we are given $a=[5, 2, 4, 3, 1]$ as input
        $pi_0=2$ because $a[0]=5$ and $a[2]=4$.
\end{itemize}

The main baseline we compare against is a NAR architecture with the same
processor trained in the no-hint regime as in \citet{bevilacqua2023neural},
however, we also provide a comparison with a more expressive processor. Unless
specified, DEARs employ the Anderson root-finding method with default
parameters from the \texttt{torchdeq} library and include Jacobian
regularization \citep{bai2021stabilizing}. Standard deviations are
based on 3 seeds.

\footnotetext[7]{As reported in \citet{bevilacqua2023neural}. Note that the processor in their experiments is Triplet-MPNN \citep{dudzik2022graph}.}
\footnotetext[8]{Run with our framework (processor is PGN; Eqs (\ref{eq:processor_start})-(\ref{eq:processor_end})). We account the improved shortest-path results to our framework's requirement that a pointer must be a graph edge which is a strong inductive bias for sparse graphs.}

\subsection{Results} We currently provide results for 4 algorithms:
Bellman-Ford, Floyd-Warshall, Strongly Connected Components (SCC) and Sorting
(insertion sort, in particular) \citep{CLRS}. Results are presented in
\Cref{tab:OHDEAR}. For completeness, although it uses a more expressive
processor, we also copy the results for the corresponding no-hint
implementation from \citep{bevilacqua2023neural}.

Our model performed on par with the baseline no-hint model for shortest-path
finding algorithms. It further outperformed the corresponding baseline on SCC,
even though it fell short of the Triplet-MPNN architecture. We further ran an
experiment adding the Jacobian regularisation term used for DEARs and
confirmed that results are not only due to the added regularisation.

\begin{table}
    \centering
    \caption{
        Test inference times (seconds/sample; A100 GPU) for NAR and DEAR
        {\footnotesize(rel.tol \& lower $\eps$)} . As runtimes for DEAR depend
        on model weights, we report standard deviation across three seeds.
    }\label{tab:runtimes}
    \begin{tabular}{lrr}
        \toprule
        & \multicolumn{2}{c}{Model}\\
        \cmidrule(r){2-3}
        \textbf{Algorithm} & \specialcell[c]{\textbf{NAR}\footnotemark[8]} & \specialcell[c]{\textbf{DEAR}\\ {\scriptsize(rel.tol \& lower $\eps$)}} \\
        \midrule
        Bellman-Ford $\textcolor{RedOrange}{\uparrow}$ & 0.028 & $0.054\pm0.004$\\
        Floyd-Warshall $\textcolor{OliveGreen}{\downarrow}$ & 0.211 & $0.146\pm0.009$\\
        SCC $\textcolor{OliveGreen}{\downarrow}$ & 0.999 & $0.055\pm0.003$\\
        Insertion Sort $\textcolor{OliveGreen}{\downarrow}$ & 1.458 & $0.050\pm0.002$\\
        \bottomrule
    \end{tabular}
\end{table}

What struck us the most was the fact that  on sorting we outperformed both the
baseline and the Triplet-MPNN architecture by a substantial margin. Noticing
that \texttt{torchdeq} uses absolute tolerance of $\eps=10^-3$ when deciding
when to stop the root-finding method we decided to test the resilience of the
method to the stopping criteria. The last result in \Cref{tab:OHDEAR} uses the
same setup but stops when the \emph{relative} error drops
below $\eps=0.1$. The results did not change significantly, however, the
processing speed improved dramatically, outperformed NAR processing
speeds on 3/4 algorithms (\autoref{tab:runtimes}) and achieving more than
10-fold improvement on sorting, while retaining accuracy. Since for sorting
datapoints are represented as full graphs, we conjecture that the solver learns
to perform some form of parallel sorting. Proving this however, requires
substantial amount of work and is left for the future.
{
    \small
    \bibliographystyle{ieeenat_fullname}
    \bibliography{main}
}


\end{document}